\documentclass[10pt,twocolumn,letterpaper]{article}

\usepackage{iccv}
\usepackage{times}
\usepackage{epsfig}
\usepackage{graphicx}
\usepackage{amsmath}
\usepackage{amssymb}
\usepackage{cite}
\usepackage{gensymb}
\usepackage{algorithm}
\usepackage{algorithmic}
\usepackage{booktabs}

\usepackage[breaklinks=true,bookmarks=false]{hyperref}

\iccvfinalcopy 

\def\iccvPaperID{****} 

\ificcvfinal\fi

\begin{document}

\title{Which to Match? Selecting Consistent GT-Proposal Assignment for Pedestrian Detection}

\author{Yan Luo$^{1}$,
Chongyang Zhang{\small $~^{1,2,}$}\footnote{*},
Muming Zhao$^{1}$,
Hao Zhou$^{1}$,
Jun Sun$^{1}$\\
$^{1}$Shanghai Jiao Tong University, Shanghai 200240, China\\
}

\maketitle
\ificcvfinal\thispagestyle{empty}\fi

\begin{abstract}
   Accurate pedestrian classification and localization have received considerable attention due to its wide applications such as security monitoring, autonomous driving, etc. Although pedestrian detectors have made great progress in recent years, the fixed Intersection over Union (IoU) based assignment-regression manner still limits their performance.
   Two main factors are responsible for this:
   1) the IoU threshold faces a dilemma that a lower one will result in more false positives, while a higher one will filter out the matched positives;
   2) the IoU-based GT-Proposal assignment suffers from the inconsistent supervision problem that spatially adjacent proposals with similar features are assigned to different ground-truth boxes, which means some very similar proposals may be forced to regress towards different targets, and thus confuses the bounding-box regression when predicting the location results.
   In this paper, we first put forward the question that \textbf{Regression Direction} would affect the performance for pedestrian detection.
   Consequently, we address the weakness of IoU by introducing one geometric sensitive search algorithm as a new assignment and regression metric. Different from the previous IoU-based \textbf{one-to-one} assignment manner of one proposal to one ground-truth box, the proposed method attempts to seek a reasonable matching between the sets of proposals and ground-truth boxes. Specifically, we boost the MR-FPPI under R$_{75}$ by 8.8\% on Citypersons dataset. Furthermore, by incorporating this method as a metric into the state-of-the-art pedestrian detectors, we show a consistent improvement.
\end{abstract}

\section{Introduction}
\begin{figure}[t]
\begin{center}
   \includegraphics[width=0.9\linewidth]{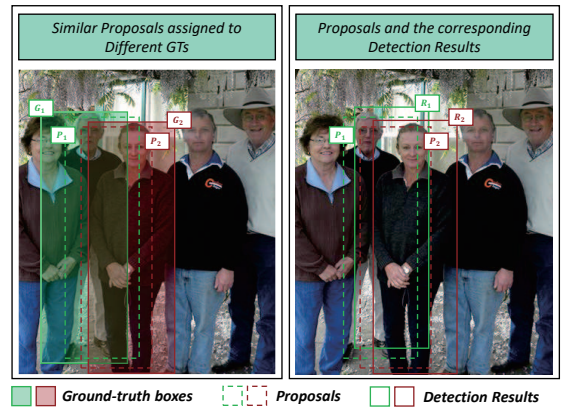}
\end{center}
   \caption{IoU-based GT-Proposal assignment may result in inconsistent regression supervision: the two proposals $\mathbf{P}_1$ and $\mathbf{P}_2$ (in the left picture), which have close location, area, and similar features, are assigned to two much different ground-truth boxes: $\mathbf{G}_1$ and $\mathbf{G}_2$, respectively. This inconsistent supervision may confuse the regression training, and thus one low-quality regression box, $\mathbf{R}_1$ in the right picture, will be predicted. With the proposed DRNet, $\mathbf{P}_1$ will be assigned to $\mathbf{G}_2$ due to less depth cost, and then one correct predicted box, $\mathbf{R}_2$ in the right, will be got.}
\label{fig:motivation}
\end{figure}
Pedestrian detection is a key problem in a number of real-world applications including auto-driving systems and surveillance systems, and is required to have both high classification and localization accuracy. Driven by the success of general object detection, many of the proposed pedestrian detectors~\cite{Zhang2018Attention, Liu2018ALF, MGAN2019, PBM2020} still follow the basic practice such as Faster R-CNN~\cite{Ren2017Faster} and SSD~\cite{Liu2015SSD}. Therefore, Intersection over Union (IoU), which is required to assign proposals, has been one indispensable metric in the recent frameworks.

However, the commonly used IoU metric has two main drawbacks. (1) On one hand, it is difficult to set a proper threshold for IoU during training. A lower IoU threshold (e.g. 0.5) could keep adequate number of positive samples but will result in many "\textit{close but not correct}" false positives during inference~\cite{Liu2018ALF}. A relative higher threshold (e.g. 0.7) can reject low-quality proposals but will reduce a large number of matched positives. Although the Cascade R-CNN~\cite{cai18cascadercnn} and ALFNet~\cite{Liu2018ALF} have provide solutions that gradually refine the proposals by several stages with a set of IoU thresholds, the hand-crafted IoU threshold, whether single or multiple, is still not the best choice.
(2) On the other hand, the IoU-based GT-Proposal assignment suffers from the inconsistent supervision problem that spatially adjacent proposals with similar features are assigned to different ground-truth boxes, which means some very similar proposals may be forced to regress towards different targets, and thus confuses the bounding-box regression when predicting the location results. This disadvantage is more prominent in the scene with large scale variation and occlusion.
As illustrated in Figure~\ref{fig:motivation}, overlapped proposals with almost the same features are assigned to different ground-truth boxes, which results in the confusion during training and thus brings about low-quality regression results during inference (Figure~\ref{fig:motivation} red solid box).

The above analyses motivate us to address the weakness of IoU-based GT-Proposal matching mechanism and propose one new assignment and regression metric for pedestrian detection. Different from the previous \textit{one-to-one} assignment manner of one proposal to one ground-truth box, we conduct the assignment in a \textit{set-to-set} process of finding a reasonable \textit{matching} between the two sets (proposals and ground-truth boxes). This searching-based method does not depend on a fixed hyper-parameter (e.g. IoU threshold), instead it constantly searches for the matched pairs along the dynamic training procedure. Furthermore, the search algorithm consists of one cost function to answer the question of "\textbf{which to match}" for each proposal. Motivated by the discovery that the distribution of pedestrian scales is highly connected with the depth-variant direction, the introduced cost function starts with the depth estimation, and gradually assigns all proposals step by step, pushing more proposals closer to ground-truth boxes with smaller depth (scale) variance. By this way, the regressor will be optimized towards consistent location direction: regress to the direction with smaller depth variance. On top of this, we are the first to point out that \textbf{"Regression needs Direction"} to the best of our knowledge, and a novel pedestrian detection architecture is thus constructed, denoted as \textbf{D}irectional \textbf{R}egression \textbf{Net}work (\textbf{DRNet}).

To demonstrate the generality of the proposed depth-guided search algorithm, we evaluate various CNN-based pedestrian detectors on both Citypersons~\cite{Zhang2017CityPersons} and Caltech~\cite{Dollar2009Pedestrian} dataset including OR-CNN~\cite{Zhang2018Occlusion}, PBM~\cite{PBM2020} and CSP\cite{liu2018CSP}.To sum up, the main contributions of this work lie in:

(1) We first attempt to put forward the \textbf{inconsistent supervision} problem of IoU-based assignment mechanism and propose one search algorithm as a new assignment metric, which is both dynamic and direction-sensitive during training procedure.

(2) We first attempt to point out that \textbf{"Regression needs Direction"} and propose one directional regression network, named DRNet, to tackle with the challenging problem of large scale variation for accurate pedestrian localization.

(3) We achieve state-of-the-art accuracy on widely-used datasets including Caltech and Citypersons, especially under the challenging setting of R$_{75}$ by 8.8\% on Citypersons dataset. We incorporate the proposed method into the most popular pedestrian detection algorithms such as CSP and PBM~\cite{PBM2020}, and show further performance improvement.

\begin{figure*}[t]
\begin{center}
   \includegraphics[width=0.85\linewidth]{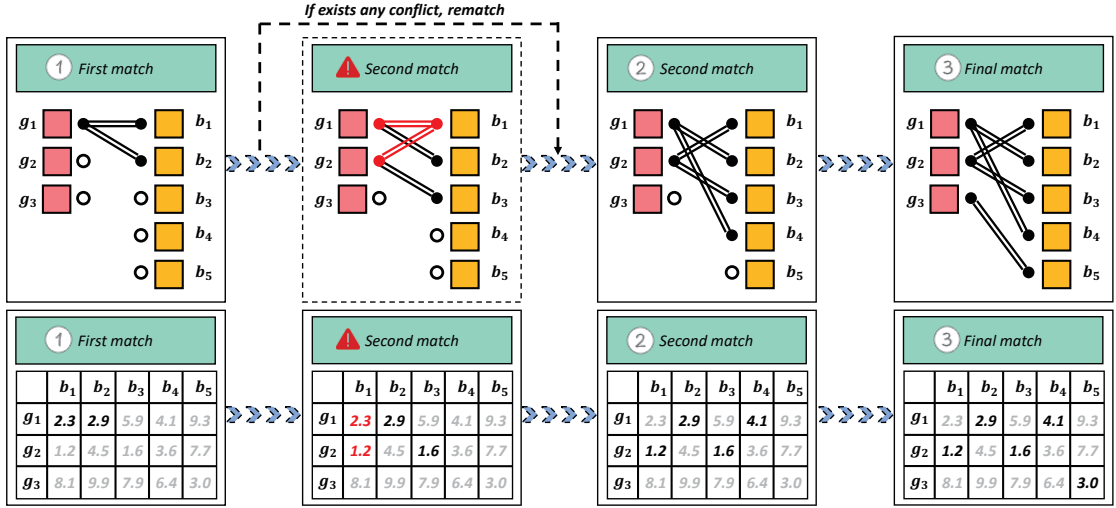}
\end{center}
   \caption{An example of the matching process of the proposed method. The second row shows matching cost of different proposal to GTs. In each step, the ground-truth box is assigned with several proposals with minimum matching cost. Besides, if exists any conflict, which means the proposal has already assigned to one ground-truth box, the matching process would be rerun. Detailed procedure is shown in Algorithm~\ref{alg:matching}.}
\label{fig:search}
\end{figure*}
\section{Related Work}
\subsection{Pedestrian Detection}
Driven by the success of general object detection, many pedestrian detectors follow the anchor-based/anchor-free paradigm. Specifically, the anchor-based detectors are proposed in the two-stage/one-stage framework.
RPN+BF~\cite{Zhang2016RPNBF} discusses the issue that improves performance of Faster R-CNN for pedestrian detection with a RPN followed by shared boosted forests.
MS-CNN~\cite{Cai2016Multiscale} attempts to introduce various receptive fields that match different object scales to tackle with multi-scale problem in pedestrian detection.
Recently, methods such as PBM~\cite{PBM2020} use both part and full-body features to more accurately classify and localize pedestrians especially in crowded scenes.
However, aforementioned detectors focus more on feature representation, and less attention is paid to the inherent drawbacks of IoU-based assignment method during the training procedure.

\subsection{Increment-based Label Assignment}
Most recently, similar to our intuition, some researchers attempt to revisit the role of IoU for object detection. First, methods such as Cascade R-CNN~\cite{cai18cascadercnn} and ALFNet~\cite{Liu2018ALF} try to replace the single threshold with an incremental IoU setting. Although these detectors have provided solutions that gradually refine the proposals, they are still not free from the hand-crafted setting and perform somewhat inflexible during training. Most recently, a progressive network~\cite{Progressive2020ECCV} is proposed with three-phase progression to gradually refine anchors following human annotation process. However, the problem is also obvious. On the one hand, although the gradual strategy could bring performance improvements, it also brings double the computational cost. On the other hand, some incremental settings are not flexible. For example, Cascade R-CNN and ALFNet both use manually defined IoU (e.g, 0.5, 0.6 and 0.7) to measure the quality.

\subsection{Statistics-based Label Assignment}
Different from previous methods that choose a fixed number of best scoring anchors, statistics-based label assignment detectors model the anchor selection procedure as a likelihood estimation for a probability distribution. ATSS~\cite{zhang2019bridging} proposes an adaptive training sample selection method that uses the sum of mean and standard deviation as the IoU Threshold. PAA~\cite{kang2020PAA} proposes a probabilistic anchor assignment strategy that adaptively separates a set of anchors into positive and negative samples for a GT box according to the learning status of the model associated with it. These methods have taken a step towards more accurate assignment strategies. Nevertheless, to the best of our knowledge, these still get caught into the IoU-based assignment,  which means they still not escape from the inconsistent problem our work raised.

\subsection{Prior-based Label Assignment}
To relief the extra computation cost, some works introduce human prior to improve label assignment. FreeAnchor~\cite{zhang2019freeanchor} proposes one detector that guarantees at least one anchor's prediction close to the ground-truth for each object. GuideAnchoring~\cite{wang2019region} and MetaAnchor~\cite{Tong2018MetaAnchor} try to dynamically change the shape of anchors to fit various distributions of objects. The aforementioned methods, whether prior-related (e.g., GuideAnchoring) or instance-related (e.g., FreeAnchor), ignore the inconsistent supervision problem existing during the assignment process.  Therefore, they are still not free from the confusion caused by "\textbf{similar proposals, different regression targets}", and thus bring about false regression results during inference. Compared to their, ours is not only dynamic and flexible, but also ensures the consistency of matching, so it better answers the question of "\textbf{which to match}" and thus shows further improvement for accurate pedestrian localization. On top of above, combined with depth estimation~(z axis), our work is also the first attempt that tries to solve the assignment problem from a 3D perspective compared with other 2D assignment (x-y axis), e.g. ATSS~\cite{zhang2019bridging} or PAA~\cite{kang2020PAA}.

\section{Proposed Method}
Generally speaking, the proposed method consists of one search algorithm and one depth-guided function to solve the main task: "\textbf{which to match}" for each proposal. The overall process is shown in Figure~\ref{fig:search}.

\subsection{How to Design Depth-guided Assignment?}
\label{sec:part2}
"\textbf{Which to Match}" is an important question especially for pedestrian detection. It is essentially a matching problem, that is how to match reasonable ground-truth boxes for each proposal so as to push the network training under a better direction. From the above analysis, the IoU metric performs as one-to-one assignment manner of one proposal to one ground-truth, but brings about inconsistent supervision problem during training. Different from IoU, we regard the assignment as the process of finding a reasonable matching between two sets (proposals and ground-truth boxes). We formulate this process in the following. The ground-truth boxes and proposals are denoted as $\mathbf{G}=\left \{ g_i \right \}_{i=1}^{M}$ and $\mathbf{B}=\left \{ b_j \right \}_{j=1}^{N}$, in which $M$ and $N$ represent the number of ground-truth boxes and proposals, respectively.

\begin{figure}[t]
\begin{center}
   \includegraphics[width=0.55\linewidth]{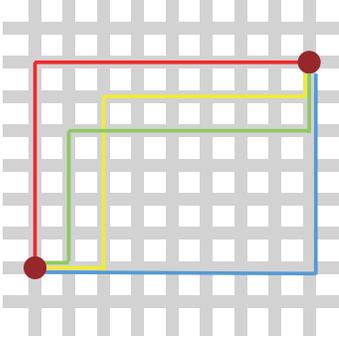}
\end{center}
   \caption{The example of the same shortest path length but with several possible paths. The red, yellow, green and blue paths all have the same shortest Manhatten length of 14. The two dark red points represent the start and end, respectively.}
\label{fig:paths}
\end{figure}
To compute the relevance between one proposal and one ground-truth box, the proposed method defines one search function, which can be divided into two parts. The first one is Manhattan distance, denoted as $\mathbf{D} (g_i, b_j)$, which measures the distance between the paired proposal and ground-truth box. It is worth noting that in the view of paths, there exist several shortest paths with the smallest Manhattan distance, as is shown in Figure~\ref{fig:paths}. The red, green, yellow and blue paths all have the same shortest path length, and this also brings some uncertainty to our second part of depth estimation, denoted as $\mathbf{Z} (g_i, b_j)$. From the start of one proposal to the end of one ground-truth box, Manhattan distance proposes several possible paths, denoted as $\mathbf{P}=\left \{ p_k \right \}_{k=1}^{T}$, in which $p_k$ represents the k-th path and T is the total number of paths. Various paths in $\mathbf{P}$ shows various depth variance, denoted as $\mathbf{V}=\left \{ v_k \right \}_{k=1}^{T}$, in which $v_{k}$ is the sum of the depth changes along the k-th path, formulated as:
\begin{equation}
    v_k=\mathbf{A}(p_k)=\sum_{q=1}^{s}d_q
\end{equation}
where $\mathbf{A}(\cdot)$ represents the aggregation function to calculate the depth variance, $s$ represents the total number of points included in the path $p_k$, and $d_q$ represents the q-th depth variance between two adjacent points. We select the path with the minimal depth variance among different paths to calculate the depth variance cost $\mathbf{Z}(g_j,b_i)$ along the matching process. It canbe defined as the following:
\begin{equation}
    \mathbf{Z}(b_i,g_j)=\mathop{\min}\ (\mathbf{V})
\end{equation}
 Consider the example in Figure~\ref{fig:search}, which schematically shows a detector with five generated proposals and three ground-truth boxes, the overall matching cost $\tilde{l}_{mc}(g_i, b_j)$ can be formulated as:
\begin{equation}
    \tilde{l}_{mc}(g_i, b_j)=\mathbf{D}(g_i, b_j)+\mathbf{Z}(g_i, b_j)
\end{equation}

\begin{algorithm}[tb]
\caption{Search Algorithm}
\label{alg:matching}
\hspace*{0.02in} {\bf Input:}\\
\hspace*{0.22in} $\mathbf{B}$ is a set of ranked proposals generated by RPN\\
\hspace*{0.22in} $\mathbf{G}$ is a set of ground-truth boxes on the image\\
\hspace*{0.22in} $M$ and $N$ are the number of GTs and proposals\\
\hspace*{0.22in} $N_p$ and $N_e$ are the number of positives and negatives\\
\hspace*{0.22in} $\tilde{l}_{mc}(g_i,b_j)$ is the matching cost with $g_i$ and $b_{j}$.\\
\hspace*{0.02in} {\bf Output:}\\
\hspace*{0.22in} $\mathcal{P}$ and $\mathcal{N}$ are sets of positives and negatives\\
\begin{algorithmic}[1] 
\STATE build an empty set for positive samples: $\mathcal{P}\leftarrow \varnothing $;
\STATE build an empty set for negative samples: $\mathcal{N}\leftarrow \varnothing $;
\STATE build an empty set for pending proposals: $\mathcal{T}\leftarrow \varnothing $;
\FOR{each box $g_i\in \mathbf{G}$} 
    \STATE $\left \langle g_i,b_j \right \rangle\leftarrow$select the proposal $b_j\in \mathbf{B}$ with the minimal cost $\tilde{l}_{mc}(g_i,b_j)$ as the assigned pair;
    \IF{the number of assigned pairs of $g_i$ $\leqslant [\frac{N_p}{M}]$}
        \STATE $\mathcal{P}= \mathcal{P}\cup \left \{ b_j \right \}$;
    \ELSE
        \STATE re-compare the cost $\tilde{l}_{mc}$ of all proposals that have assigned to $g_i$;
        \IF{the proposal with the largest cost is $b_j$}
            \STATE $\mathcal{T}= \mathcal{T}\cup \left \{ b_j \right \}$;
        \ELSE
            \STATE $b_l \leftarrow$ proposal with the largest cost;
            \STATE $\mathcal{P}= \mathcal{P}\cup \left \{ b_j \right \}$;$\mathcal{P}=\mathcal{P}-\left \{ b_l \right \} \leftarrow$ abandon $b_l$;
            \STATE $\mathcal{T}=\mathcal{T}\cup \left \{ b_l \right \} \leftarrow$ put it into the pending set;
        \ENDIF
    \ENDIF
    \STATE $\mathbf{B}=\mathbf{B}-\left \{ b_j \right \}$;
\ENDFOR
\IF{$(|\mathcal{N}|+|\mathcal{T}|)<N_e$}
    \STATE compute IoU between $\mathbf{G}$ and $\mathbf{B}$;
    \STATE randomly assign negative samples to fill $\mathcal{N}$;
\ENDIF
\RETURN $\mathcal{P}$, $\mathcal{N}$, $\mathcal{T}$;
\end{algorithmic}
\end{algorithm}

where $\mathbf{D}$ and $\mathbf{Z}$ represent the value of Manhatten distance and depth variance of all proposals. In the following, we consider a simple assignment strategy based on the estimated matching cost $l_{mc}(g_i, b_j)$. We sort the proposals with the corresponding confidence estimated by Region Proposal Network (RPN). The proposed matching strategy sequentially assigns proposals to the ground-truth boxes. In details, we define the maximum number of assignment of each ground-truth box as $N_p/M$, in which $N_p$ represents the number of positive samples and $M$ represents the number of ground-truth boxes of the input image. For each proposal in $\mathbf{B}=\left \{ b_j \right \}_{j=1}^{N}$, we sequentially assign the ground-truth box with the smallest $l_{mc}$ during its matching. Detailed matching process is shown in Algorithm~\ref{alg:matching}.

It is worth noting that
(1) each ground-truth boxes can assign up to $[N_p/M]$ proposals; (2) the proposal that was previously assigned to ground-truth box but was removed latter is sampled as hard negative samples; (3) if the negative samples are still not full at the end, IoU is also introduced to fill the negative set. Besides, in order to enrich the diversity of negative samples, half of negative proposals are randomly sampled from the negative set $\mathcal{N}$, and the remaining half are from pending set $\mathcal{T}$. The matching algorithm assigns a unique proposal to each ground-truth box. Given this algorithm, we define a loss function on pairs of sets $\mathbf{G}$ and $\mathbf{B}$ as:
\begin{equation}
\begin{aligned}
    L(\mathbf{G},\mathbf{B})&=\alpha \sum_{j=1}^{N_p}l_{reg}(b_{j},\tilde{g})+\beta \sum _{j=1}^{N_p+N_e}l_{cls}(b_j,\tilde{g})
\end{aligned}
\label{eq:loss}
\end{equation}
where $l_{reg}$ is a regression loss, $l_{mc}$ is the matching cost and $l_{cls}$ is a cross-entropy loss on a proposal's confidence that it would be matched to a ground-truth box. The label for this cross-entropy loss is provided by $\tilde{g}$. $\alpha$, $\beta$ and $\eta$ are the parameters trading off between the three parts. Note that for the proposed matching, we can update the network by back-propagating the gradient of this loss function.
\subsection{Implement Details}
\label{sec:part3}
The above algorithm attempts to answer the question "\textbf{which to match}" for each proposal. In this section, we hope to further improve the proposed algorithm. We focus on the most important two aspects: (1) how to assign proposals with various scales to different ground-truth boxes? (2) how to relief inconsistency between depth-guided assignment and IoU-based performance evaluation?

To answer the first question, we follow the practice in FCOS~\cite{FCOS2019}, which introduces multi-level prediction with FPN~\cite{lin2016fpn}. With the use of multi-level feature maps defined as $\left \{ C_3,C_4,C_5, C_6,C_7 \right \}$, proposals and ground-truth boxes with different scale would be allocated to the specific feature level. The specific feature level is only used to assign proposals with the specific scale. During experiments, the size range is (0, 64] for $C_3$, (64, 128] for $C_4$, (128, 256] for $C_5$, (256, 512] for $C_6$ and (512, $+\infty$] for $C_7$. Since proposals with different sizes are allocated to different feature levels, the proposals and ground-truth boxes on the same level would be closer with larger overlap, which further enhances our algorithm.

In order to ensure our consistency between training and evaluation, we attempt to incorporate assignment into the confidence score of each predicted bounding box. We add one parallel branch to the detection head of classification and regression to predict the matching cost ${l}_{mc}$ of each proposal. The aforementioned calculated matching cost $\tilde{l}_{mc}$ servers as the estimation target to supervise the whole process. Therefore, the total loss in Equation~\ref{eq:loss}, redefined as:
\begin{equation}
        \tilde{L}(\mathbf{G},\mathbf{B})=L(\mathbf{G},\mathbf{B})+\gamma \sum_{j=1}^{N_p}\mathbf{Logistic}(l_{j, mc}, \tilde{l}_{j, mc})
\end{equation}
where Logistic($\cdot$) means the loss function using in standard Logistic Regression. In this way, we try to reduce the confidence of some detected boxes, which regress different from our matching procedure. In details, taking the proposal $b_t$ as an example, the regressed box of $b_t$ is denoted as $\tilde{b}_{t}$. The actual matching cost between $b_t$ and $\tilde{b}_t$ can be calculated following Equation~\ref{eq:loss}, denoted as $l^{*}_{mc}(b_{t}, \tilde{b}_{t})$. The confidence score $s_t$ of $b_t$ is re-measured, formulated as:
\begin{equation}
   \tilde{s}_{t} = s_t \cdot [1.5-\mathbf{Sigmoid}(|l^{*}_{mc}- l_{mc}|)]
\end{equation}

\begin{table*}
\centering
\caption{Comparisons with the state-of-the-art methods on Citypersons validation dataset. Results are the MR$^{-2}$ evaluation metric of the corresponding methods, in which lower is better. \textbf{Boldface} indicates the \textbf{best} performance. \textit{ex.} means whether the method introduces extra annotations. For example, PBM and PRNet use extra part annotation, and TLL+MRF considers the sequence input.}
\begin{tabular}{ccccccccc}
\toprule[2pt]
\textbf{Method}&\textbf{Year}&\textit{ex.}&\textbf{Backbone}&\textbf{Scale}&\textbf{R}$_{50}$&\textbf{R}$_{75}$&\textbf{Heavy}&\textbf{Inference Time}\\
\midrule[1pt]
Faster R-CNN+ATT~\cite{Zhang2018Attention}&CVPR2018&$\checkmark$&VGG-16&$\times \ 1.0$&16.0&48.1&56.7&-\\
TLL~\cite{TLL2018}&ECCV2018&$\times$&ResNet-50&$\times \ 1.0$&15.5&44.2&53.6&-\\
TLL+MRF~\cite{TLL2018}&ECCV2018&$\checkmark$&ResNet-50&$\times \ 1.0$&14.4&-&52.0&-\\
ALFNet~\cite{Liu2018ALF}&ECCV2018&$\times$&ResNet-50&$\times \ 1.0$&12.0&36.5&51.9&0.27s\ /\ img\\
RepLoss~\cite{Wang2017Repulsion}&CVPR2018&$\times$&ResNet-50&$\times \ 1.3$&11.6&-&55.3&-\\
OR-CNN~\cite{Zhang2018Occlusion}&ECCV2018&$\times$&VGG-16&$\times \ 1.0$&12.8&-&55.7&-\\
MGAN~\cite{MGAN2019}&ICCV2019&$\times$&VGG-16&$\times \ 1.0$&11.5&-&51.7&-\\
PBM~\cite{PBM2020}&CVPR2020&$\checkmark$&VGG-16&$\times \ 1.0$&11.0&-&53.3&-\\
CSP~\cite{liu2018CSP}&CVPR2019&$\times$&ResNet-50&$\times \ 1.0$&11.0&34.7&49.3&0.33s\ /\ img\\
PRNet~\cite{Progressive2020ECCV}&ECCV2020&$\checkmark$&ResNet-50&$\times \ 1.0$&10.8&-&53.3&0.22s\ /\ img\\
\midrule[1pt]
DRNet (ours)&-&$\times$&VGG-16&$\times \ 1.0$&10.2&29.7&46.9&\textbf{0.12s\ /\ img}\\
&&&&$\times \ 1.3$&9.5&27.0&43.6&0.14s\ /\ img\\
DRNet (ours)&-&$\times$&ResNet-50&$\times \ 1.0$&10.1&25.9&46.2&0.15s\ /\ img\\
&&&&$\times \ 1.3$&\textbf{9.1}&\textbf{25.4}&\textbf{42.1}&0.18s\ /\ img\\
\bottomrule[2pt]
\end{tabular}
\label{tab:citypersons}
\end{table*}

\section{Experiments}
We assess the effectiveness of our proposed method for pedestrian detection on widely used datasets Caltech~\cite{Doll2012Pedestrian}~\cite{Dollar2009Pedestrian} and Citypersons~\cite{Zhang2017CityPersons}.

\subsection{Experimental Setup}
The proposed method is based on the Faster R-CNN baseline~\cite{Ren2017Faster}, pre-trained on the ImageNet. We optimize the network using the Stochastic Gradient Descent (SGD) algorithm with 0.9 momentum and 0.0005 weight decay, which is trained on 1 1080Ti GPU with the mini-batch involving 1 image per GPU. For Caltech dataset, we train the network for $40k$ iterations with the initial learning rate of $10^{-3}$ and decay it to $10^{-4}$ for another $20k$ iterations. For Citypersons dataset, we train the network for $20k$ iterations with the initial learning rate of $10^{-3}$ and decay it to $10^{-4}$ for another $10k$ iterations. All images are in the original $1\mathrm{x}$. scale during training and testing. Other parameters $\alpha$, $\beta$ and $\gamma$ are set to 1, 1 and 0.01, respectively. $N_p$ and $N_e$ are both equal to 256 as usual. Specifically, our depth estimation model is VNL~\cite{Yin2019enforcing}. Better depth estimation model may have better performance, but it is not in the main scope of ours method.

\subsection{Evaluation Metrics}
In experiments, we used the standard \textit{average-log miss rate} (MR) on \textit{False Positive Per Image} (FPPI) in [$10^{-2}$, $10^0$]. This kind of metric is a bit similar to \textit{Average Precision} (MAP) and refers more to the object not detected.

On Caltech and Citypersons, we report results across different occlusion degrees: \textbf{Reasonable}, \textbf{Heavy} and \textbf{Partial}. The visibility ratio is (0.65, 1], (0, 0.65) and (0.65, 1), respectively. In all subsets, the height of pedestrian over 50 pixels is taken for evaluation. It is worth noting that \textbf{Heavy} is designed to evaluate performance in case of severe occlusions. To further demonstrate our performance, we have designed two special settings. The first one is to validate the localization accuracy. We not only test on Reasonable set under IoU$=0.5$ ($\mathbf{R}_{50}$), but also on Reasonable set under IoU$=0.75$ ($\mathbf{R}_{75}$). The second one is that we split the Citypersons into a new subset called Large Height Variation (\textbf{LHV}), in which each image contains height variation of pedestrian instances larger than 50 pixels with some occlusion, in which the visibility ratio is [0.2, 0.9].

\begin{table*}
\centering
\caption{Comparisons with different modules on Cityperons validation dataset. Results are the MR$^{-2}$ evaluation metric of the corresponding methods, in which lower is better. \textbf{Boldface} indicates the \textbf{best} performance. Directional Assignment is described in Sec~\ref{sec:part2} and Refinement is described in Sec~\ref{sec:part3}}
\begin{tabular}{ccccccccc}
\toprule[2pt]
\textbf{Backbone}&\textbf{Scale}&\textbf{IoU Assignment}&\textbf{Direction Assignment}&\textbf{Refinement}&\textbf{R}$_{50}$&\textbf{Heavy}&\textbf{Partial}&\textbf{Bare}\\
\midrule[1pt]
VGG-16&$\times \ 1.0$&$\checkmark$&&&15.8&53.2&16.9&9.7\\
VGG-16&$\times \ 1.0$&&$\checkmark$&&11.8&49.6&15.5&9.3\\
VGG-16&$\times \ 1.0$&&$\checkmark$&$\checkmark$&\textbf{10.2}&\textbf{46.9}&\textbf{14.5}&\textbf{6.6}\\
\bottomrule[2pt]
\end{tabular}
\label{tab:ablation}
\end{table*}

\begin{figure*}[htpb]
\begin{center}
   \includegraphics[width=0.86\linewidth]{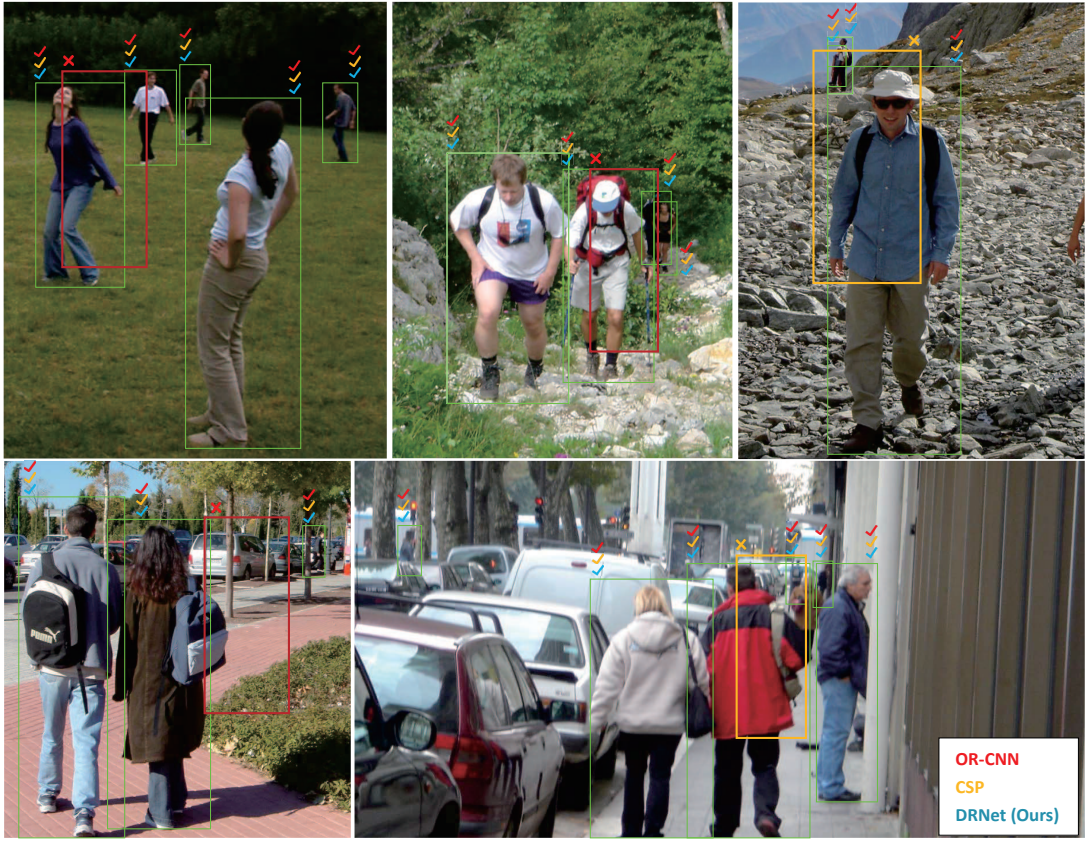}
\end{center}
   \caption{The visualization results of our DRNet and the state-of-the-art methods, e.g., CSP~\cite{liu2018CSP} and OR-CNN~\cite{Zhang2018Occlusion}. The box with the check mark or cross represents the correct and false results, respectively. The red and yellow cross show false positives of OR-CNN and CSP, while ours performs better.}
\label{fig:visual}
\end{figure*}

\subsection{Main Results}
We compare DRNet with corresponding methods on Citypersons and Caltech dataset in Table~\ref{tab:citypersons} and Table~\ref{tab:Caltech}. For fair comparisons, we report results in terms of backbone, scale, inference time and all challenging subsets.

\textbf{(1) Citypersons Dataset}. Table~\ref{tab:citypersons} reports the results compared to state-of-the-arts on Citypersons. First of all, our algorithms has significant performance improvement under various fairness settings. For example, when leveraging ResNet-50, ours achieves the highest accuracy with an improvement of 0.9\% MR$^{-2}$ from the closet competitor CSP~\cite{liu2018CSP} on \textbf{R}$_{50}$ and 8.8\% MR$^{-2}$ on \textbf{R}$_{75}$. Furthermore, it is worth noting that from Table~\ref{tab:citypersons}, ours has also demonstrated the self-contained ability to handle occlusion issues in crowded scenes. Especially on \textbf{Heavy} occlusion subset, ours reports the brand-new state-of-the-arts of 42.1\% MR$^{-2}$. This is probably because harder samples are mined with the introduced directional assignment metric, and thus training a more discriminant predictor. And we mainly improve the training matching process, there is no additional modification to the network structure. Therefore, there is no extra computation consumption during testing, and our inference time is comparable to others.

\textbf{(2) Caltech Dataset}. We also test our method on Caltech and the comparison with state-of-the-arts on this benchmark is shown in Table~\ref{tab:Caltech}. Our method achieves MR$^{-2}$ of 3.08\% under the IoU threshold of 0.5, which is comparable to the best competitor (3.80\% of CSP~\cite{liu2018CSP}). Besides, in the case of a stricter occlusion level of Heavy subset, our method achieves 30.45\% MR$^{-2}$, outperforming all previous state-of-the-arts with an improvement of 6.05\% MR$^{-2}$ over CSP~\cite{liu2018CSP}. It indicates that our method has a substantially better localization accuracy.

\subsection{Ablation}
To show the effectiveness of each proposed component, we report the overall ablation studies in Table~\ref{tab:ablation}.

\textbf{(1) Search and Assignment}. As analyzed above, it can be seen that the IoU metric are suboptimal primarily because it is difficult to answer "\textbf{which to match}" question for each proposal. The performance is summarized in Table~\ref{tab:ablation}. When evaluated respectively, the search and assignment strategy shows the improvement of 4.0\% MR$^{-2}$ compared with the original IoU assignment. Furthermore, by incorporating with the refinement module, we show a consistent improvement of 1.6\% MR$^{-2}$. In addition, we can also see that from the visualization results in Figure~\ref{fig:visual}, our algorithm effectively reduces false detection results when there are large height variance of pedestrians in each image compared with other algorithms, e.g., CSP~\cite{liu2018CSP} or OR-CNN~\cite{Zhang2018Occlusion}. Taking the first image in Figure~\ref{fig:visual} as an example, the red dotted box failed to regress toward the green box (boy with white T-shirt), and thus brings about "close but not correct" false positives. Moreover, our algorithm is particularly effective in two challenging scenarios. One is the scene with large height variance, such as the example in Figure~\ref{fig:visual} (first row, third column). The confusion in IoU makes it harder to train a high quality regressor, and thus brings about "close but not correct" false positives during testing, while ours can effectively eliminate this kind of low-quality regression. The second is the occlusion scene. In this case, our algorithm has more consistent matching targets and therefore has better performance (Figure~\ref{fig:visual}, second row, second column).

\begin{table}
\centering
\caption{Comparisons with the state-of-the-art methods on Citypersons validation dataset. \textbf{Boldface} indicates the \textbf{best} performance. $\dagger$ means our re-implementation of referred method, and $\dagger \dagger$ means the corresponding detector combined with our assignment manner.}
\begin{tabular}{cccc}
\toprule[2pt]
\textbf{Method}&\textbf{Backbone}&\textbf{R}$_{50}$&\textbf{LHV}\\
\midrule[1pt]
BiBox~\cite{Zhou2018Bi}&VGG-16&11.2&-\\
BiBox$^\dagger$&VGG-16&11.2&30.6\\
BiBox$^{\dagger \dagger}$&VGG-16&\textbf{10.5}&\textbf{24.7}\\
\midrule[1pt]
PBM~\cite{PBM2020}&VGG-16&11.1&-\\
PBM$^\dagger$&VGG-16&11.0&30.3\\
PBM$^{\dagger \dagger}$&VGG-16&\textbf{10.2}&\textbf{23.9}\\
\midrule[1pt]
CSP~\cite{liu2018CSP}&ResNet-50&11.0&-\\
CSP$^\dagger$&ResNet-50&10.9&29.0\\
CSP$^{\dagger \dagger}$&ResNet-50&\textbf{10.2}&\textbf{23.7}\\
\bottomrule[2pt]
\end{tabular}
\label{tab:ablationHV}
\end{table}

\begin{table}
\centering
\caption{Comparisons with the state-of-the-art methods on Caltech testset. \textbf{Boldface} indicates the \textbf{best} performance.}
\begin{tabular}{cccc}
\toprule[2pt]
\textbf{Method}&\textbf{Backbone}&\textbf{R}$_{50}$&\textbf{Heavy}\\
\midrule[1pt]
DeepParts~\cite{Liu2017Deep}&-&12.90&60.42\\
MS-CNN~\cite{Cai2016Multiscale}&-&8.08&59.94\\
RPN+BF~\cite{Zhang2016RPNBF}&-&7.28&74.36\\
SDS-RCNN~\cite{SDSRCNN2017}&ResNet-50&6.44&58.55\\
ALFNet~\cite{Liu2018ALF}&ResNet-50&4.50&43.40\\
RepLoss&ResNet-50&4.00&63.36\\
CSP~\cite{liu2018CSP}&ResNet-50&3.80&36.50\\
\midrule[1pt]
DRNet (ours)&ResNet-50&\textbf{3.08}&\textbf{30.45}\\
\bottomrule[2pt]
\end{tabular}
\label{tab:Caltech}
\end{table}

\begin{figure}[htpb]
\begin{center}
   \includegraphics[width=0.8\linewidth]{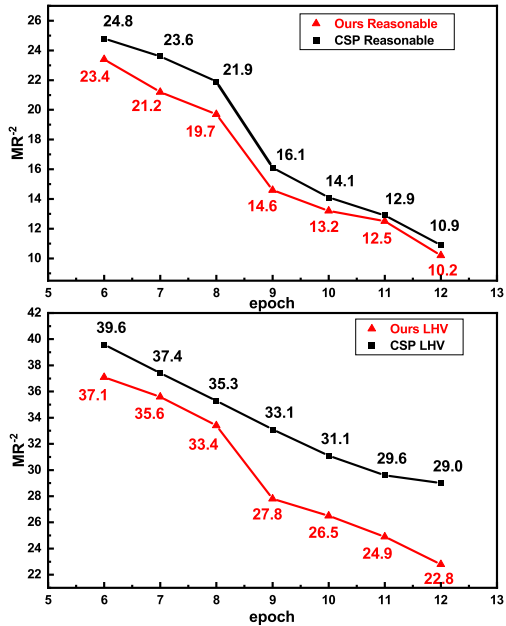}
\end{center}
   \caption{The performance of MR$^{-2}$ along training process under different epoch. \textbf{LHV} means the maximum pedestrian height variance per image.}
\label{fig:AP0.5&0.75}
\end{figure}

\textbf{(2) Large Height Variance}. In order to better verify our algorithm, we specially set a \textbf{L}arge \textbf{H}eight \textbf{V}ariance subset (\textbf{LHV}). The separation of this subset is based on the maximum variance in pedestrian height per image. On one hand, in Figure~\ref{fig:AP0.5&0.75}, we show that the change of miss rate during training process under different settings. We found that our method can accurately detect under large height variance (\textbf{LHV}). On the other hand, we also integrate the algorithm with other algorithms and verify its performance on the \textbf{LHV} subset (Table~\ref{tab:ablationHV}). In order to make a fair comparison, we reproduced the relevant algorithm and report a comparison between our reproduced performance and the performance reported in corresponding paper. When integrated with other algorithms, the performance has also been greatly improved, especially on the \textbf{LHV} subset. However, we have also noticed that although our algorithm has improved different detectors (e.g., CSP or PBM), results report in Table~\ref{tab:ablationHV} just show competitive with the bare search algorithm in Table~\ref{tab:citypersons}. We think this is because there exist some incompatibility between the previous algorithms and ours method. For example, PBM~\cite{PBM2020} and BiBox~\cite{Zhou2018Bi} both introduce part annotations, which leads to some confusion in our assignment procedure.

\begin{table}
\centering
\caption{Comparisons with the state-of-the-art methods on COCO minival set.  $\dagger$ means the corresponding detector combined with our assignment manner. \textbf{Boldface} indicates the \textbf{best} performance.}
\begin{tabular}{cccc}
\toprule[2pt]
\textbf{Method}&\textbf{Backbone}&\textbf{AP}&\textbf{AP}$_{75}$\\
\midrule[1pt]
FPN~\cite{lin2016fpn}&ResNet50&33.9&-\\
RetinaNet~\cite{Focal2017}&ResNet50&36.3&38.8\\
FCOS~\cite{FCOS2019}&ResNet50&36.6&38.9\\
FreeAnchor~\cite{zhang2019freeanchor}&ResNet50&38.7&41.6\\
ATSS~\cite{zhang2019bridging}&ResNet50&39.3&42.8\\
PAA~\cite{kang2020PAA}&ResNet50&41.1&44.3\\
\midrule[1pt]
FPN$^\dagger$&ResNet50&37.5&40.9\\
FCOS$^\dagger$&ResNet50&41.4&44.7\\
\textbf{RetinaNet}$^\dagger$&\textbf{ResNet50}&\textbf{41.5}&\textbf{45.0}\\
\bottomrule[2pt]
\end{tabular}
\label{tab:COCO}
\end{table}

\textbf{(2) Extended Experiments.} In order to further validate our algorithm, we also report results on COCO minival dataset compared with other detectors. We use a COCO training setting which is the same as~\cite{kang2020PAA} in the batch size, frozen Batch Normalization, learning rate, etc. For ablation studies, we use ResNet-50 backbone and run 135 iterations of training. We notice that the performance of the proposed label assignment is improved by adding it to other widely-used frameworks, such as 3.6\% on FPN~\cite{lin2016fpn}, 5.2\% on RetinaNet~\cite{Focal2017} and 4.8\% on FCOS~\cite{FCOS2019}. Furthermore, compared with other label assignment strategies, ours also show competitive performance in Table~\ref{tab:COCO}. Although our algorithm is designed specially for pedestrian detection, it still has some performance improvement for more general detection dataset such as COCO. We believe that the performance will be further improved if there is a more refined design for the general purpose.

\section{Conclusion}
In this paper, we present a simple yet effective pedestrian detector with a novel assignment strategy, achieving competitive accuracy while performing competitive inference time with the state-of-the-art methods. On top of a backbone, the proposed method can serve as a metric incorporated into the other pedestrian detectors, and experimental results show a consistent improvement on some popular pedestrian detection benchmarks, e.g. Citypersons and Caltech. This novel design is flexible and independent of any backbone network, without being limited by the two-stage detection framework. Therefore, it is also interesting to incorporate the proposed assignment strategy with other detectors like FCOS~\cite{FCOS2019} and YOLO~\cite{Redmon2017YOLO9000}, which will be studied in future. Furthermore, we also consider extending this method, not only for pedestrian detection, but also study the proposed method on more general object detection tasks.

{\small
\bibliographystyle{ieee_fullname}
\bibliography{egbib}
}

\end{document}